\def\year{2019}\relax
\documentclass[letterpaper]{article} % DO NOT CHANGE THIS
\usepackage{aaai19}  % DO NOT CHANGE THIS
\usepackage{times}  % DO NOT CHANGE THIS
\usepackage{helvet} % DO NOT CHANGE THIS
\usepackage{courier}  % DO NOT CHANGE THIS
\usepackage[hyphens]{url}  % DO NOT CHANGE THIS
\usepackage{graphicx} % DO NOT CHANGE THIS
\usepackage{graphicx}  % DO NOT CHANGE THIS
\usepackage{graphicx}
\usepackage{longtable}
\usepackage{xspace}
\usepackage{multirow}
\usepackage{tabularx}

\newcommand{\whatif}{``what if" local explainability\xspace}

\title{Assessing the Local Interpretability of Machine Learning Models}
\author{ \ \Large \textbf{Dylan Slack,\textsuperscript{\rm 1} Sorelle A. Friedler,\textsuperscript{\rm 1}, Carlos Scheidegger\textsuperscript{\rm 2}, Chitradeep Dutta Roy}\textsuperscript{\rm 3}\\ % All authors must be in the same font size and format. Use \Large and \textbf to achieve this result when breaking a line
\textsuperscript{\rm 1}Haverford College\\ %If you have multiple authors and multiple affiliations
\textsuperscript{\rm 2}University of Arizona\\
\textsuperscript{\rm 3}University of Utah\\
}
 
\begin{document}

\maketitle

\begin{abstract}The increasing adoption of machine learning tools has led to calls for accountability via model interpretability.
  But what does it mean for a machine learning model to be interpretable by humans, and how can this be assessed?
  We focus on two definitions of interpretability that have been introduced in the machine learning literature: simulatability (a user's ability to run a model on a given input) and \whatif (a user's ability to correctly determine a model's prediction under local changes to the input, given knowledge of the model's original prediction).
  Through a user study with $1000$ participants, we test whether humans perform well on tasks that mimic the definitions of simulatability and \whatif on models that are typically considered locally interpretable. 
  To track the relative interpretability of models, we employ a simple metric, the runtime operation count on the simulatability task.
  We find evidence that as the number of operations increases, participant accuracy on the local interpretability tasks decreases.
  In addition, this evidence is consistent with the common intuition that decision trees and logistic regression models are interpretable and are more interpretable than neural networks.
  
%\keywords{Interpretable machine learning; Explainable artificial intelligence; Transparency}
\end{abstract}

\section{Introduction}

Recently, there has been growing interest in interpreting machine learning models.
The goal of interpretable machine learning is to allow oversight and understanding of machine-learned decisions.
Much of the work in interpretable machine learning has come in the form of devising methods to better explain the predictions of machine learning models.
However, such work usually leaves a noticeable gap in understanding interpretability~\cite{lipton2018mythos,doshi2017towards}.
The field currently stands on shaky foundations: papers mean different things when they use the word ``interpretability'', and interpretability claims are typically not validated by measuring human performance on a controlled task.  However, there is growing recognition in the merit of such human validated assessments \cite{lage2018humaninloop,lage2018evaluation,lakkaraju2016}.
In line with this goal, we seek concrete, \emph{falsifiable} notions of interpretability.

``Interpretability'' can be broadly divided into \emph{global interpretability}, meaning understanding the entirety of a trained model including all decision paths, and \emph{local interpretability}, the goal of understanding the results of a trained model on a specific input and small deviations from that input.
We focus on local interpretability, and on two specific definitions.
We assess \emph{simulatability} \cite{lipton2018mythos}, the ability of a person to---independently of a computer---run a model and get the correct output for a given input, and \emph{\whatif} \cite{ribeiro2016should,lipton2018mythos}: the ability of a person to correctly determine how small changes to a given input affect the model output.
We will refer to a model as \emph{locally interpretable} if users are able to correctly perform \emph{both} of these tasks when given a model and input.
The experiments we present here are necessarily artificial and limited in scope.
We see these as lower bounds on the local interpretability of a model; if people cannot perform these interpretability tasks, these models should not be deemed locally interpretable.

In addition to considering the successful completion of these tasks a lower bound on the local interpretability of a model, we might reasonably ask whether these are valuable interpretability tasks at all.
Though purposefully limited in scope, we argue that these tasks are still valuable in real-world settings.
Consider a defense attorney faced with a client's resulting score generated by a machine learned risk assessment.
In order to properly defend their client, the attorney may want to verify that the risk score was correctly calculated (simulatability) and argue about the extent to which small changes in features about their client could change the calculated score (local explainability).
Despite being simple interpretability tasks, successfully completing them is important to the attorney's ability to defend their client from potential errors or issues with the risk assessment. 

%
%Consider a defense attorney attempting to verify a risk assessment score.   The defense attorney could be interested in how the algorithm produced a certain risk score and how small changes to the input affect that risk score.  Depending on the type and size of the model, the defense attorney may exhibit varying levels of ability, if at all, to accomplish the desired goal.  It would be useful to have available a set of criteria that specifically rule out the interpretability of the model to the defense attorney.

We assessed the simulatability and \whatif of decision trees, logistic regressions, and neural networks through a crowdsourced user study using Prolific~\cite{prolific}.  We asked 1,000 participants to simulate the model on a given input and anticipate the outcome on a slightly modified version of the input.
We measured user accuracy and completion time over varied datasets, inputs, and model types (described in detail in the User Study Design section).
The results are consistent with the folk hypotheses \cite{lipton2018mythos} that decision trees and logistic regression models are locally interpretable and are more locally interpretable than neural networks given the particular model representations, datasets, and user inputs used in the study.

As has been previously observed \cite{lipton2018mythos}, it may be the case that a small neural network is more interpretable than a very large decision tree.
To begin to answer questions surrounding cross-model comparisons and generalizations of these results to models not studied here, we investigated a measure for its suitability as a proxy for the users' ability to correctly perform both the simulation and \whatif tasks.
We hypothesized that the number of program operations performed by an execution trace of the model on a given input would be a good proxy for the time and accuracy of users' attempts to locally interpret the model under both definitions; specifically, that as the total number of operations increased, the time taken would increase and the accuracy on the combined task would decrease.

Analyzing the results of this study, we find evidence that as the number of total operations performed by the model increases, the time taken by the user increases and their accuracy on the combined local interpretability task decreases.
We anticipated that as the number of operations increases, the model would become uninterpretable because all users are eventually expected to make a mistake simulating a very large model.
The operation count at which the users cannot locally interpret a model can be considered an upper bound limit to the interpretability of the model.
Users reached this upper bound when simulating the largest neural network sizes we considered.
We see this work as a first step in a more nuanced understanding of the users' experience of interpretable machine learning.

\section{Related Work}

Work on the human interpretability of machine learning models began as early as Breiman's study of random forests \cite{breiman2001random}.
Since then, many approaches to the interpretability of machine learning models have been considered, including the development of new globally interpretable models \cite{ustun2016supersparse}, post-hoc local explanations \cite{ribeiro2016should} and visualizations \cite{olah2018the}, and post-hoc measurement of the global importance of different features \cite{henelius2014peek,datta2016algorithmic,adler2018auditing}.
We refer the interested read to Molnar and Guidotti et al. for a more detailed discussion of these methods~\cite{molnar2018interpretable,guidotti2018survey}.

Some of the recent activity on interpretability has been prompted by Europe's General Data Protection Regulation (GDPR).
A legal discussion of the meaning of the regulation with respect to interpretability is ongoing.
Initially, the GDPR regulations were described as providing a ``right to an explanation''~\cite{goodman2016european}, although subsequent work challenges that claim \cite{wachter2017right}, supporting a more nuanced right to ``meaningful information'' about any automated decision impacting a user~\cite{selbst2017meaningful}.
Exactly what is meant by interpretability to support the GDPR and in a broader legal context remains in active discussion~\cite{selbst2018intuitive}.

The uncertainty around the meaning of ``interpretability'' has prompted calls for more precise definitions and carefully delineated goals~\cite{lipton2018mythos}.
One thought-provoking paper makes the case for a research agenda in interpretability driven by user studies and formalized metrics that can serve as validated proxies for user understanding~\cite{doshi2017towards}.
Doshi-Velez and Kim argue that human evaluation of the interpretability of a method in its specific application context is the pinnacle of an interpretability research hierarchy followed by human evaluation of interpretability on a simplified or synthetic task and analysis of proxy tasks without associated user studies.
In order to perform interpretability analysis without user studies, they argue, it is necessary to first assess proxies for user behavior.
Here, we propose one such metric and assess its suitability as a proxy for the local interpretability of a model.

Although we are unaware of existing metrics for the local interpretability of a \emph{general} model, 
%there are a number of existing metrics for global interpretability of models.  The \emph{model description length} connects the choice of model and parameters to information theory by defining the best model as the one with the minimal compressed size \cite{rissanen1978modeling}.  There are also 
many measures developed by the program analysis community aim at assessing the understandability of a general \emph{program}, which could be seen as metrics for global interpretability.
For example, the \emph{cyclomatic complexity} counts the number of independent paths through a program using its control flow graph~\cite{mccabe1976complexity}.  Metrics for \emph{specific} model types have also been developed.  Lage et al. \cite{lage2018evaluation} investigate how different measures of complexity in decision sets affect accuracy and response time on tasks consisting of simulatability, verification, and counterfactual-reasoning.  Via six different user studies of 150 people (for a total of 900 participants) they find that increased complexity in decision set logic results in increased response time but do not find a significant connection with accuracy.  They measure decision set complexity as a combination between the explanation size, clauses in the disjunctive normal form of the input (called cognitive chunks), and number of repeated input conditions to the decision set.  Their work is specific to decision sets and does not generalize to other model types.
%

%propose a method to optimize the interpretability of decision trees for a variety of suggested interpretability metrics.  Because they consider only decision tree interpretability metrics, their findings regarding the usefulness of their metrics are not model agnostic.  

There have also been experimentally grounded assessments of model properties related to (but different from) interpretability.  
%Despite calls for more experimentally grounded assessments of interpretability \cite{doshi2017towards,abdul2018trends}, there have so far been few user studies focusing on the interpretability of machine learning models.  
Poursabzi-Sangdeh et al. \cite{poursabzi2017manipulating} consider the impact of model attributes (e.g. black-box vs. clear) on user 
%measure how the number of features and model transparency affect
 trust, simulatability, and mistake detection using randomized user studies on a similar scale to what we will consider here.  %While we focus on assessing model types, Poursabzi-Sangdeh et al. vary across general model attributes (e.g. black box vs. clear).  
 They find that clear models (models where the inner calculations are displayed to the user) are best simulated.
Allahyari et al. \cite{Allahyari2011understandability} measure the \emph{perceived} relative understandability of decision trees and rule-based models and find decision trees are seen as more understandable than rule-based models.  

Other methods are concerned with human in the loop optimization of the interpretability of machine learning models. Lage et. al. \cite{lage2018humaninloop} develop a method that optimizes models for both interpretability and accuracy by including user studies in the optimization loop.  Their method minimizes the number of user studies needed to generate models that are both interpretable and accurate.  They perform experiments on optimizing decision trees and find that the proxy interpretability metric optimized by the model (e.g. number of nodes, mean path length) varies based on dataset.

%Veale et al. \cite{veale2018fairness} surveyed 27 public sector machine learning professionals to determine current challenges facing the successful implementation of their work.
%They call for a greater focus on transparent machine learning in order to bridge the gap between those implementing algorithms and those overseeing the direction of public projects.
%Particularly, they argue that machine learning algorithms applied to public settings sit in an intersection between partisan politics, aging infrastructure, and institution aided lock-in.
%Thus, they recommend researchers strive for solutions that provide a high level of understanding to researchers but also those managers and bureaucrats charged with their supervision.
%Considering the growing need for transparency, 
%We believe that systems that make it possible to compare the relative interpretability of models could assist researchers and professionals in making more appropriate choices for their particular contexts.
%end comment

\section{A Metric for Local Interpretability}

\begin{figure}[!ht]
\begin{center}
\includegraphics[width=\columnwidth]{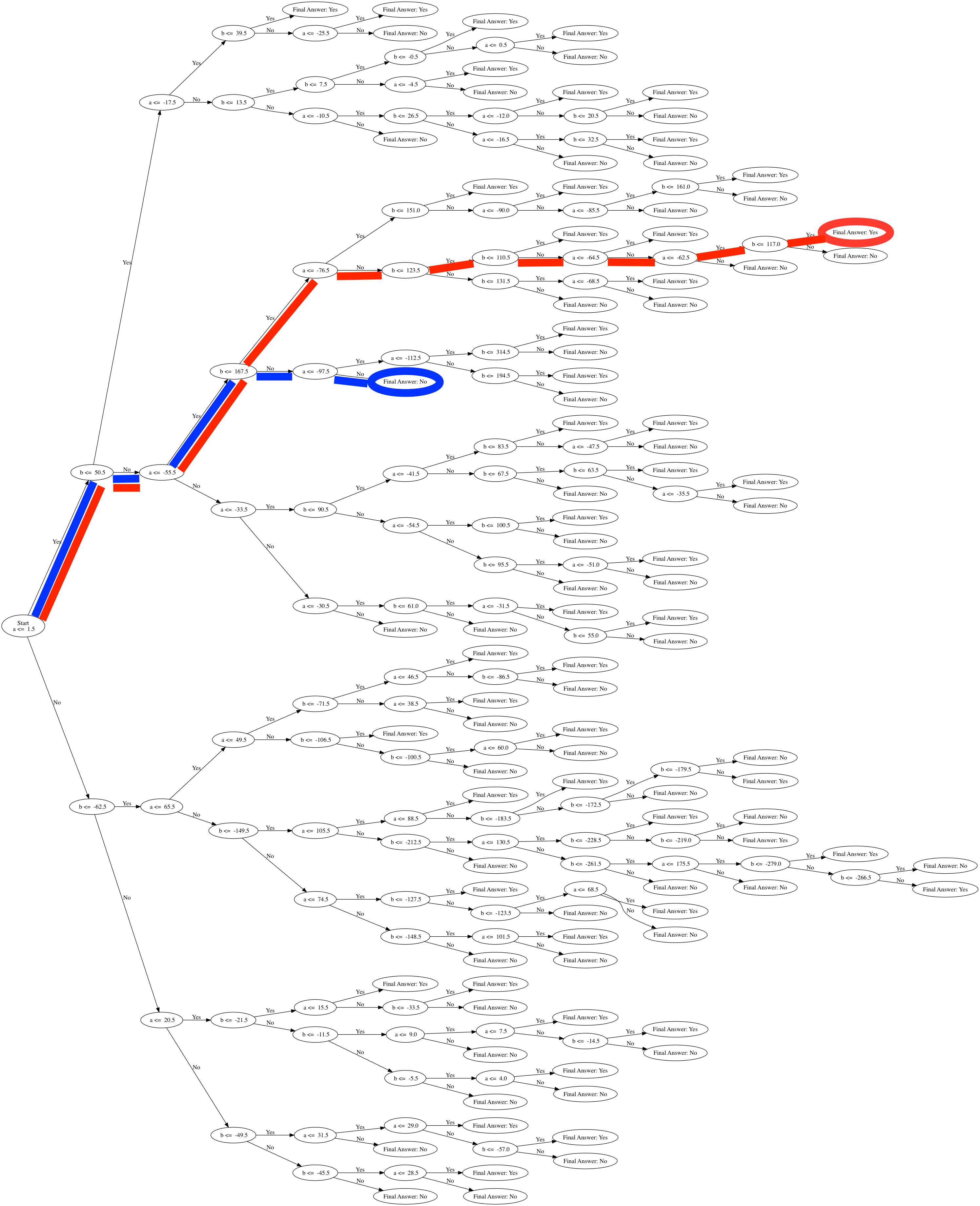}
\caption{A decision tree where the answer when run on the input $(a = -80, b = 200)$ is shown circled in blue and the result of running the same model on the input $(a = -64, b = 115)$ is shown circled in red.}
\label{fig:dt2d}
\end{center}
\end{figure}

Motivated by the previous literature and its calls for user-validated metrics that capture aspects of interpretability, we wish to assess whether a candidate metric captures a user's ability to simulate \emph{and} ``what if" locally explain a model.
The candidate metric we consider here is the \emph{total number of runtime operation counts} performed by the model when run on a given input.
We consider two basic variants of operations, arithmetic and boolean, and track their totals separately.
Effectively, we seek a proxy for the work that a user must do (in their head or via a calculator) in order to simulate a model on a given input, and will claim that the total number of operations also impacts a user's ability to perform a ``what if" local explanation of a model.
%Even though this is clearly a basic approximation of the process performed by users, 
%we note that even this simple model appears to not have been studied explicitly.

\subsection{An Example}
As an example of how this metric would work, consider the visualization of a decision tree in Figure~\ref{fig:dt2d}.
The result of running the model on the input $(a = -80, b = 200)$ is shown circled in blue and the result of running the same model on the input $(a = -64, b = 115)$ is shown circled in red.
The red answer is at a depth of 10 in the decision tree while the blue answer is at a depth of 5.
Counting the operations that the model takes to run on the input (including each boolean comparison operation or memory access required, which we count as an arithmetic operation) gives the total number of runtime operations - our candidate metric.
Using the below methodology to count these operations, 
to determine the number of runtime operations executed when evaluating the decision tree model on the inputs from the example above, 
(blue: $a = -80, b = 200$ and red: $a = -64, b = 115$), 
the blue input is found to require 17 total operations (6 operations are arithmetic and 11 are boolean) while the red input requires 32 total operations (11 arithmetic and 21 boolean).  Essentially, at each branch point one arithmetic operation is performed to do a memory access, one boolean operation is performed to check if the node is a leaf node, and one more boolean operation is performed for the branching operation.

\subsection{Calculating Runtime Operation Counts}

In order to calculate the number of runtime operations for a given input, we instrumented the prediction operation for existing trained models in python's scikit-learn package~\cite{sklearn_api}. The source code for the technique is available at \emph{URL removed for anonymization}. %\url{https://github.com/darkreactions/measuring_interpretability}
Since most machine learning models in scikit-learn use (indirectly, via other dependencies) cython, Fortran, and C for speed and memory efficiency, we implemented a pure Python version of the \texttt{predict} method for the classifiers, and instrumented the Python bytecode directly.
We created pure-Python versions of the decision tree, logistic regression, and neural network classifiers in scikit-learn.\footnote{Specifically, \texttt{sklearn.tree.DecisionTreeClassifier}, \texttt{sklearn.linear\_model. LogisticRegression}, and \texttt{sklearn.neural\_network.MLPClassifier}.}

Once working only with pure Python code, we used the tracing feature of python's \texttt{sys} module and a custom tracer function to count the number of boolean and arithmetic operations.
The default behavior of tracer in python is line based, meaning the trace handler method is called for each line of the source code.
% But for this study we needed to know every operation at the bytecode level to determine whether it is boolean or arithmetic.
We used the \texttt{dis} module to modify the compiled bytecode objects of useful modules stored in their respective .pyc files.
In particular, we modified the line numbering metadata so that every bytecode is given a new line number, ensuring that our tracer function is called for every bytecode~instruction \cite{pychack,pycfile,vmbook}.
Inside the tracer function we use the \emph{dis} module to determine when a byte corresponds to a valid operation and count them accordingly for our simplified \emph{predict} method implementations when run on a given input.

\section{User Study Design}
\label{design}

We have two overall goals in this project: to assess the simulatability and \whatif of machine learning models, and to study the extent to which the proposed metric works as proxy for local interpretability.
To those ends, we designed a crowdsourced experiment that was given to $1000$ participants.
Participants were asked to run a model on a given input and then evaluate the same model on a locally changed version of the input.
We start by describing the many potentially interacting factors that required a careful experimental design.

\subsection{Models and Representations}

For this study we consider the local interpretability of three models: decision trees, logistic regression, and neural networks.
We chose decision trees and logistic regression because they are commonly considered to be interpretable~\cite{lipton2018mythos}.
In contrast, we picked neural networks because they are commonly considered uninterpretable.
%These models also make for interesting objects of study since they operate in fundamentally different ways.
%Decision trees rely heavily on boolean operations (branching structure), logistic regression relies heavily on arithmetic operations, and neural networks use both types of operations.
The models were trained using the standard package scikit-learn.\footnote{Decision trees were trained using \texttt{sklearn.tree.DecisionTreeClassifier} without any depth restrictions and with default parameters.
  Logistic regression was trained using \texttt{sklearn.linear\_model.LogisticRegression} with the multi\_class argument set to '\emph{multinomial}' and '\emph{sag}'(Stochastic average gradient descent) as the solver.
  The neural network was implemented using \texttt{sklearn.neural\_network.MLPClassifier}.
The neural network used is a fully connected network with 1 input layer, 1 hidden layer with 3 nodes, and 1 output layer.  The \texttt{relu} (rectified linear unit) activation function was used for the hidden layer.}

%Perhaps the largest problem in trying to assess whether a user can interpret a model is that the user needs some sort of intelligible representation of the model. While it might be reasonable to say that the code run by the model in the \texttt{predict} method is the ``purest'' representation of the model, it is not likely to be understandable by a lay audience without a programming background, and might additionally suffer from code readability issues. Thus, we sought to create representations that would reveal all of the model's calculations and choices more directly. Our goal was explicitly \emph{not} to create visualizations that might give users higher-level insights into the models' operation, but to keep to an unelaborated form of the model. Still, all our results should be understood as assessing a combination of our representations together with the models, rather than studying the models in isolation.

\begin{figure}[htb]

\begin{center}
\includegraphics[width=\columnwidth]{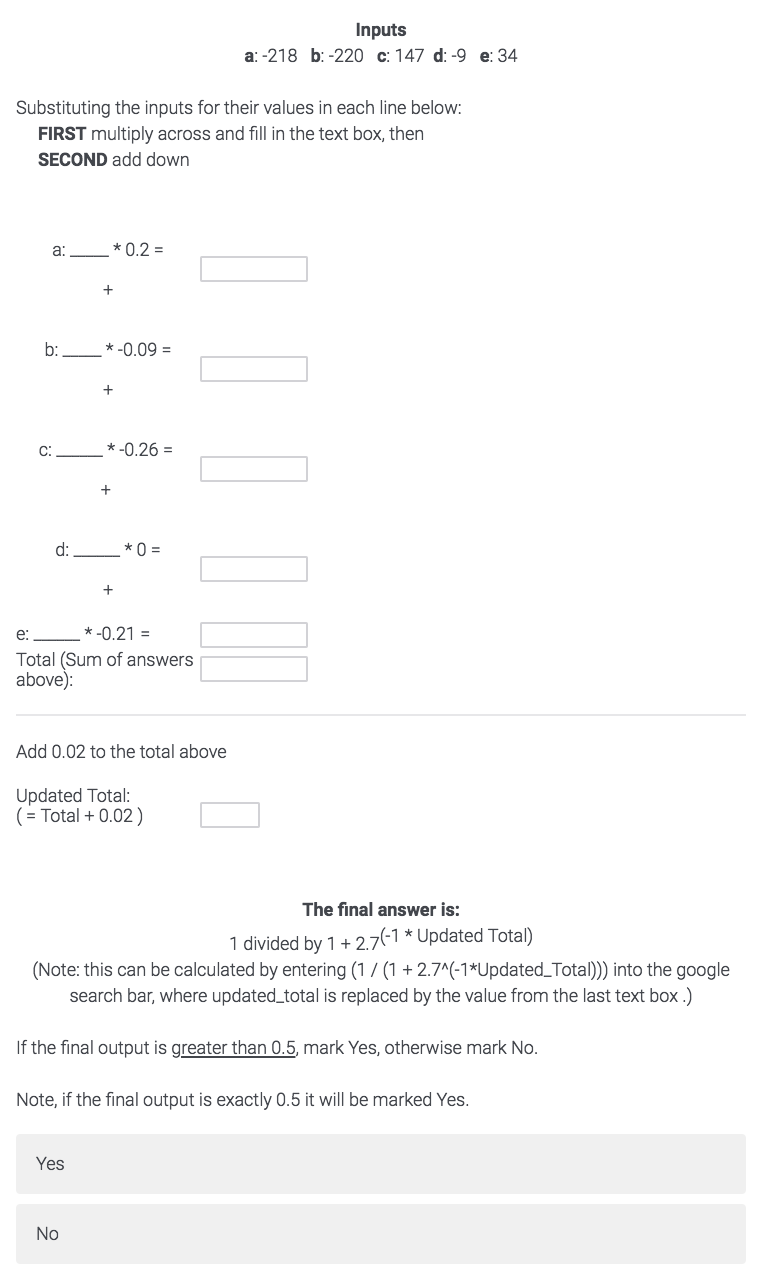}
\caption{The logistic regression representation shown to users.}
\label{fig:lr}
\end{center}
\end{figure}

Our decision tree representation is a standard node-link diagram representation for a decision tree or flow chart.
%Each node is a decision point and each edge is labeled with the result of the preceding node's decision.
%Given an input, the trained decision tree model starts at the root node and proceeds down the tree, using the given input to answer the node questions and follow the decision path.
%The leaf nodes indicate the final classification result.
In order to allow users to simulate the logistic regression and neural network classifiers we needed a representation that would walk the users through the calculations without previous training in using the model or any assumed mathematical knowledge beyond arithmetic.  The resulting representation for logistic regression is shown in Figure \ref{fig:lr}.  The neural network representation used the same representation as the logistic regression for each node and one page per layer. 

%In order to allow users to simulate the logistic regression and neural network classifiers we needed a representation that would walk the users through the calculations without previous training in using the model or any assumed mathematical knowledge beyond arithmetic.  The resulting representation for logistic regression is shown in Figure \ref{fig:lr}.  The neural network representation used the same representation as the logistic regression for each node and one page per layer.  The activation results from the previous layer were shown to the user as new inputs for the next layer.  An example of the resulting representation is shown in Figure \ref{fig:nn}.

\begin{figure}[htbp]

\begin{center}

\includegraphics[width=\columnwidth]{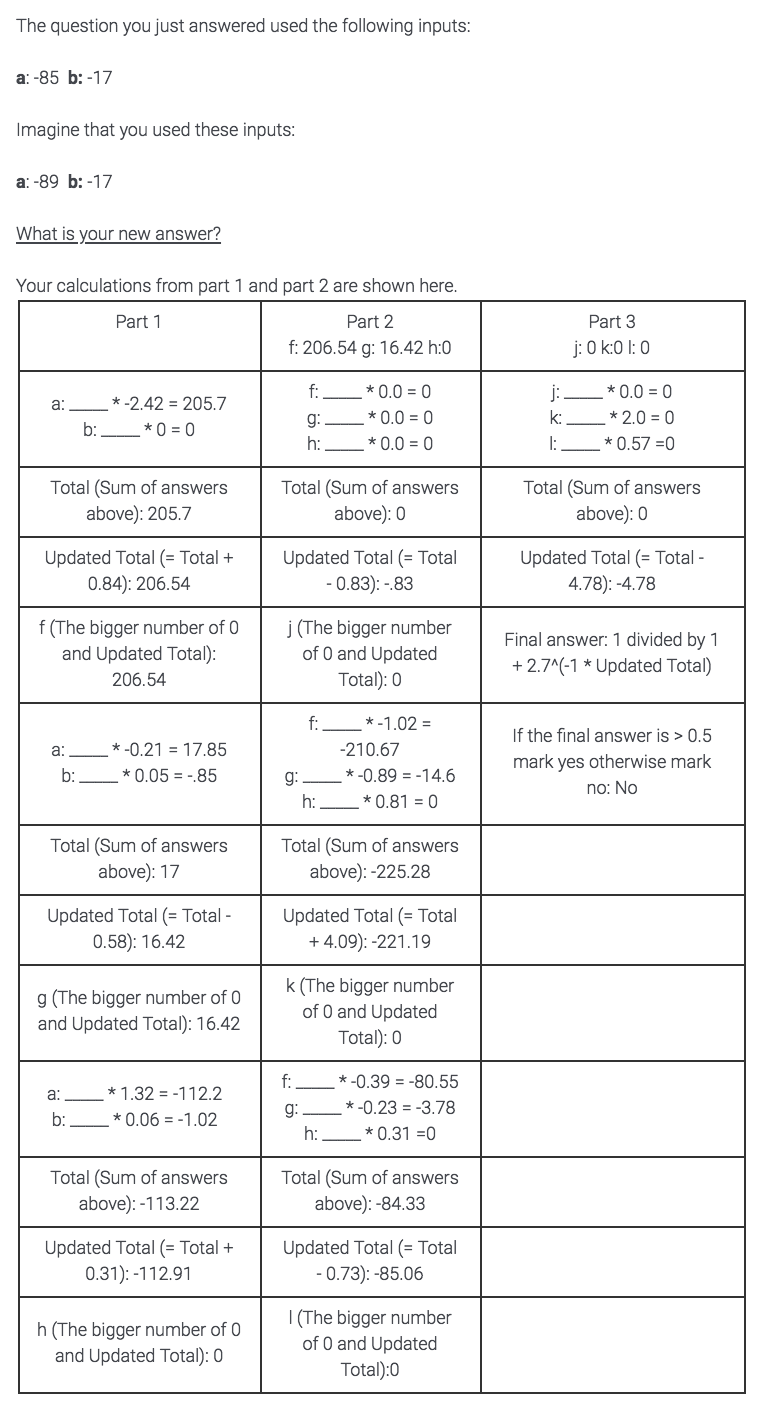}
\includegraphics[width=\columnwidth]{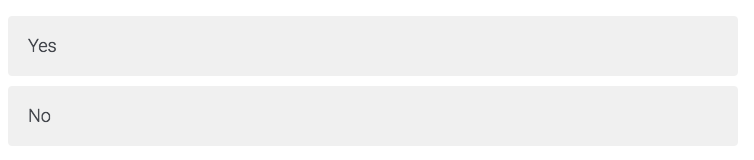}
\end{center}

\caption{The \whatif question shown to users for the neural network model.  Note that while the simulatability question on the neural networks allowed users to fill in the blanks, the shown blanks in the above image represent where the variable values will be filled in, and users are given no location to fill in partial simulations of the neural network.}
\label{fig:nn}
\end{figure}

%The representations described so far are for the first question a user will be asked about a model - the request to simulate it on a given input.  In order to allow users to assess the \whatif of the model, we will also ask them to determine the output of the model for a perturbed version of the initial input they were shown.  The representations used here are the same as the ones above, but a snapshot of the participants' previously filled in answers are shown for the logistic regression and neural network representations (see the neural network perturbed input question in Page 4 of the participant's view in Figure \ref{fig:nn}).

The representations described so far are for the first question a user will be asked about a model - the request to simulate it on a given input.  In order to allow users to assess the \whatif of the model, we also asked them to determine the output of the model for a perturbed version of the initial input they were shown.  The representations used here are the same as the ones described, but a snapshot of the participants' previously filled in answers are shown for the logistic regression and neural network representations (see Figure \ref{fig:nn}) and users are \emph{not} given blank entries to allow the re-simulation of the model.

\subsection{Data and Inputs}

\label{sec:data}

In order to avoid effects from study participants with domain knowledge, we created synthetic datasets to train the models.  We created four synthetic datasets simple enough so that each model could achieve 100\% test accuracy.  These datasets consisted of a 2 dimensional dataset with rotation around an axis applied, 2 dimensional without rotation around an axis, 3 dimensional with rotation around an axis, and 5 dimensional with rotation around an axis.  As the number of dimensions increases, so does the operation count.  These four datasets were used to train the three considered models via an 80/20 train-test split.  We generated user inputs using the test data. For each test data point, we changed one dimension incrementally in order to create a perturbed input.  

From this set of input and perturbed input pairs, we then chose a set of eight pairs for each trained model (i.e., for each model type and dataset combination) to show to the participants.  The set was chosen to fit the following conditions: 50\% of the classifications of the original inputs are True, 50\% of the classifications on the perturbed input are True, and 50\% of the time the classification between input and its perturbed input changes.  We used this criteria in order to distribute classification patterns evenly across users so that a distribution of random guesses by the participants would lead to 50\% correctness on each task, and guessing that the perturbed input had the same outcome as the original input would also be correct 50\% of the time.

\subsection{Pilot Studies}

In order to assess the length of the study and work out any problems with instructions, we conducted three pilot studies.  In the first informal study, one of us watched and took notes while a student attempted to simulate an input on each of the three types of models and determine the outcome for a perturbed input for each of those three models.  In the second two pilots we recruited about 40 participants through Prolific and gave the study for a few fixed models and inputs with the same setup as we would be using for the full study.  The main takeaways from these pilot studies were that we estimated it would take users 20-30 minutes to complete the survey, but that some users would take much longer.  We had originally planned to include a dataset with 10 dimensions, and based on the time taken by users in the pilot survey decreased our largest dataset to 5 dimensions and added the 2-dimensional dataset with no rotation.  %We also used these pilots to confirm that we were collecting all the data we would want for our future analyses and found that our Qualtrics setup was missing timing information for the perturbed input question; this was added for the full experiment.

\subsection{Experimental Setup}

We used Prolific to distribute the survey to 1000 users each of whom was paid \$3.50 for completing it.  Participants were restricted to those with at least a high school education (due to the mathematical nature of the task) and a Prolific rating greater than 75 out of 100.  The full survey information (hosted through Qualtrics) and resulting data is available online.\footnote{\emph{URL removed for anonymization}} %https://github.com/darkreactions/measuring_interpretability/

Each participant was asked to calculate the output of a machine learning model for a given input, and then to determine the output of a perturbed input applied to the same model.  We showed each participant three trained models: a logistic regression, a decision tree, and a neural network in a random order.  Each participant was shown a model trained on a specific dataset (chosen from the four described earlier) at most once to avoid memory effects across models.   Each question began with the initial input and a brief description of the task.  As an attention check, we included a question in the survey that asked users to do some basic addition.  Lastly, we asked each user at the end of the study to indicate whether they fully attempted to determine correct answers and that they would still be compensated in case they selected no.  We considered only the data of the 930 users, who we will refer to as \emph{confident respondents}, that selected they fully tried to determine correct answers and who correctly answered the basic addition problem.

\paragraph*{Preregistered Hypotheses}
We preregistered two experimental hypotheses.  Namely, that time to complete will be positively related to operation count and that accuracy will be negatively related to operation count.  We also preregistered two exploratory hypotheses.  These were that we would explore the specific relationship between time and accuracy versus operation count and that we would explore how the perturbed input is related to time and operation count.  These hypotheses can be found at the Open Science Framework at: \textit{url removed for anonymization}

\paragraph*{Study Setup Issues}
After running the user study, we found that an error in the survey setup meant that the survey exited prematurely for users given two of the eight inputs on the decision tree models for one dataset. %the 2-dimensional dataset with rotation.  
Since we did not receive data from these participants, Prolific recruited other participants who were allocated to other inputs and datasets, so the analyzed dataset does not include data for these two inputs.  Users who contacted us to let us know about the problem were still paid.

\paragraph*{Multiple Comparison Corrections} %, Confidence Intervals.} 
In order to mitigate the problem of multiple comparisons, all p-values and confidence intervals we report in the next section include a Bonferroni correction factor of 28.  While we include 15 statistical tests in this paper, we considered a total of 28. Reported p-values greater than one arise from these corrections. %All confidence intervals are reported at 95\% confidence.

\section{User Study Results}

Based on the results from the described user study, we now examine folk hypotheses regarding the local interpretability of different model types, consider the relative local interpretability of these models, and assess our proposed metric.

%relative interpretability between models as well as assess our proposed metrics.  In order to gauge the relative interpretability, we will look at the relative correct / incorrect ratios for both the initial input and the perturbed input questions and verify that our results are statistically sound through a Fisher exact test.  To assess our candidate metric of the total number of operations, we consider one regression analysis per model type and assess the significance of the metric in determining both time and accuracy.  

\subsection{Assessing the Local Interpretability of Models}

\begin{table}
\begin{center}
\begin{tabular*}{\columnwidth}{l @{\extracolsep{\fill}} lll}
				&	&	& ``What If" \\
                                     &         & Simulatability    &        Local Explainability                \\ \cline{2-4}
\multirow{3}{*}{DT}       & Correct & 717 / 930 & 719 / 930                        \\
                                     & p-Value & $5.9 \times 10^{-63}$  & $5.16 \times 10^{-64}$  \\
                                     & 95\% CI & $[0.73, 0.81]$         & $[0.73, 0.82]$                \\ \cline{2-4}
\multirow{3}{*}{LR} & Correct & 592 / 930              & 579 / 930                        \\ 
                                     & p-Value & $1.94 \times 10^{-15}$  & $2.07 \times 10^{-12}$    \\
                                     & 95\% CI & $[0.59, 0.69]$         & $[0.57, 0.67]$              \\ \cline{2-4}
\multirow{3}{*}{NN}      & Correct & 556 / 930              & 499 / 930                   \\
                                     & p-Value & $7.34 \times 5.5^{-8}$ & $0.78$                    \\
                                     & 95\% CI & $[0.55,0.65]$          & $[0.49, 0.59]$    
                                     
%                                                                          &         & Simulatability               & \whatif              & AND                    \\ \cline{2-5}
%\multirow{3}{*}{Decision Tree}       & Correct & 717 / 930              & 719 / 930              & 594 / 930              \\
%                                     & p-Value & $4.4 \times 10^{-63}$  & $3.87 \times 10^{-64}$ & $4.7,\times 10^{-16} $ \\
%                                     & 95\% CI & $[0.73, 0.81]$         & $[0.73, 0.81]$         & $[0.59, 0.69]$         \\ \cline{2-5}
%\multirow{3}{*}{Logistic Regression} & Correct & 592 / 930              & 579 / 930              & 425 / 930              \\ 
%                                     & p-Value & $1.5 \times 10^{-15}$  & $1.5 \times 10^{-12}$  & $1.6 \times 10^{-12}$  \\
%                                     & 95\% CI & $[0.59, 0.68]$         & $[0.57, 0.67]$         & $[0.57, 0.67]$         \\ \cline{2-5}
%\multirow{3}{*}{Neural Network}      & Correct & 556 / 930              & 499 / 930              & 425 / 930              \\
%                                     & p-Value & $1.5 \times 5.5^{-8}$ & $0.59$                 & $0.20$                  \\
%                                     & 95\% CI & $[0.55,0.65]$          & $[0.49, 0.59]$         & $[0.41, 0.51]$        
\end{tabular*}
\end{center}
\caption{Per-model correct responses out of the total confident respondents on the original input (simulatability task) and perturbed inputs (\whatif task) for decision trees, logistic regression, and neural networks.  $p$-values given are with respect to the null hypothesis that respondents are correct 50\% of the time, using exact binomial tests.}
\label{tab:permodel}
\end{table}

In order to assess the local interpretability of different model types, we first separately consider the user success on the task for simulatability (the original input) and the task for \whatif (the perturbed input).  Since inputs were chosen so that 50\% of the correct model outputs were ``yes'' and 50\% were ``no'', we compare the resulting participant correctness rates to the null hypothesis that respondents are correct 50\% of the time.  The resulting $p$-values and confidence intervals are shown in Table \ref{tab:permodel}.

The results indicate strong support for the simulatability of decision trees, logistic regression, and neural networks based on the representations the users were given.  The results also indicate strong support for the \whatif of decision trees and logistic regression models, but neural networks were \emph{not} found to be ``what if" locally explainable.

Recall that we consider models to be locally interpretable if they are \emph{both} simulatable and ``what if" locally explainable.  Based on the results in Table \ref{tab:permodel}, we thus have evidence that decision trees and logistic regression models are locally interpretable and neural networks are not, partially validating the folk hypotheses about the interpretability of these models.  Next, we'll consider the relative local interpretability of these models.

\subsection{Assessing Relative Local Interpretability}

\begin{table*}
\label{table:fisherexact}
\caption{Comparative correct / incorrect distributions and $p$-values between model types generated through Fisher Exact Tests for confident responses.  Relative correctness is shown for simulatability (correctness on the original input), \whatif (correctness on the perturbed input), and local interpretability (correctness on both parts).  DT stands for Decision Tree, LR stands for Logistic Regression, and NN stands for Neural Network.}
\begin{center}
\begin{center}
\textbf{Relative Simulatability:}\end{center}
\begin{tabular}{lllllll}
\\
 Contingency Table & \multicolumn{2}{c}{DT $>$ NN} & \multicolumn{2}{c}{DT $>$ LR} & \multicolumn{2}{c}{LR $>$ NN} \\ 
Correct   & 717 & 556   & 717 & 592   & 592 & 556 \\
Incorrect & 213 & 374   & 213 & 338   & 338 & 374 \\
p-value, 95\% CI & $1.5\times 10^{-14}$ & $[1.69, \infty]$ & $3.7\times 10^{-9}$  & $[1.43, \infty]$ & $1.3 \nobreakspace\nobreakspace\nobreakspace\nobreakspace\nobreakspace\nobreakspace\nobreakspace\nobreakspace\nobreakspace\nobreakspace $ & $[0.90, \infty]$ \\
\end{tabular}

\begin{center}\textbf{Relative ``What If" Local Explainability:}\end{center}
\begin{tabular}{lllllll}

 Contingency Table & \multicolumn{2}{c}{DT $>$ NN} & \multicolumn{2}{c}{DT $>$ LR} & \multicolumn{2}{c}{LR $>$ NN} \\ 
Correct   & 719 & 499   & 719 & 579   & 579 & 499 \\
Incorrect & 211 & 431   & 211 & 351   & 351 & 431 \\
p-value, 95\% CI & $7.3\times 10^{-26}$ & $[2.20, \infty]$& $2.6 \times 10^{-11}$ &$[1.54, \infty]$& $2.9 \times 10^{-3} $& $[1.09, \infty]$ \\
\end{tabular}

\begin{center}\textbf{Relative Local Interpretability:}\end{center}
\begin{tabular}{lllllll}

 Contingency Table & \multicolumn{2}{c}{DT $>$ NN} & \multicolumn{2}{c}{DT $>$ LR} & \multicolumn{2}{c}{LR $>$ NN} \\ 
Correct   & 594 & 337   & 594 & 425   & 425 & 337 \\
Incorrect & 336 & 593   & 336 & 505   & 505 & 593 \\
p-value, 95\% CI & $9.3\times 10^{-32}$ & $[2.36, \infty]$& $5.9 \times 10^{-14}$ &$[1.60, \infty]$& $5.7 \times 10^{-4} $& $[1.13, \infty]$ \\
\end{tabular}

%\begin{center}\textbf{Results for the perturbed input conditioned a correct response on the original input:}\end{center}
%\begin{tabular}{ccccccc}
%
% Contingency Table & \multicolumn{2}{c}{Decision Tree > Neural Net} & \multicolumn{2}{c}{Decision Tree > Logistic Regression} & \multicolumn{2}{c}{Logistic Regression > Neural Net} \\ 
% Correct   & 594 & 337   & 594 & 425   & 425 & 337 \\
% Incorrect & 123 & 219   & 123 & 167   & 167 & 219 \\
% p-value,  95\% CI & $2.15\times 10^{-16}$ & $[2.15, \infty]$& $1.18 \times 10^{-6}$  & $[1.29, \infty]$&$ 3.95 \times 10^{-5}$& $[1.15, \infty]$ \\
%\end{tabular}

\end{center}
\label{tab:comparemodel}
\end{table*}

In order to assess the relative local interpretability of models --- to evaluate the folk hypothesis that decision trees and logistic regression models are more interpretable than neural networks --- we compared the distributions of correct and incorrect answers on both tasks across pairs of model types.  We applied one-sided Fisher exact tests with the null hypothesis that the models were equally simulatable, ``what if" locally explainable, or locally interpretable.  The alternative hypotheses were that decision trees and logistic regression models were more interpretable (had a greater number of correct responses) than neural networks and that decision trees were more interpretable than logistic regression.

The results (see Table \ref{tab:comparemodel}) give strong evidence that decision trees are more locally interpretable than logistic regression or neural network models on both the simulatability and \whatif tasks.  While there was strong evidence that logistic regression is more ``what if" locally explainable and more locally interpretable than neural networks, there is not evidence that logistic regression is more simulatable than neural networks using the given representations.  This may be because the logistic regression and neural network representations were very similar.  An analysis of the users who got both tasks right, i.e., were able to locally interpret the model, shows that the alternative hypothesis was strongly supported in all three cases, thus supporting the folk hypotheses that decision trees and logistic regression models are more interpretable than neural networks.

\subsection{Assessing Runtime Operations as a Metric for Local Interpretability}

%\begin{figure}[htbp]
%\begin{center}
%\includegraphics[width=2in]{figures/totalopsVtime_nopresentationerrors.png} 
%\includegraphics[width=2in]{figures/arithmeticopsVtime_nopresentationerrors.png}
%\caption{The number of operations (top: total, bottom: arithmetic operations only) taken to calculate the model's outcome on a specific input versus the time taken by confident users to simulate the model on the given input.}
%\label{fig:opsVtime}
%\end{center}
%\end{figure}

\begin{figure*}[!ht]
\begin{center}
\includegraphics[scale = .14]{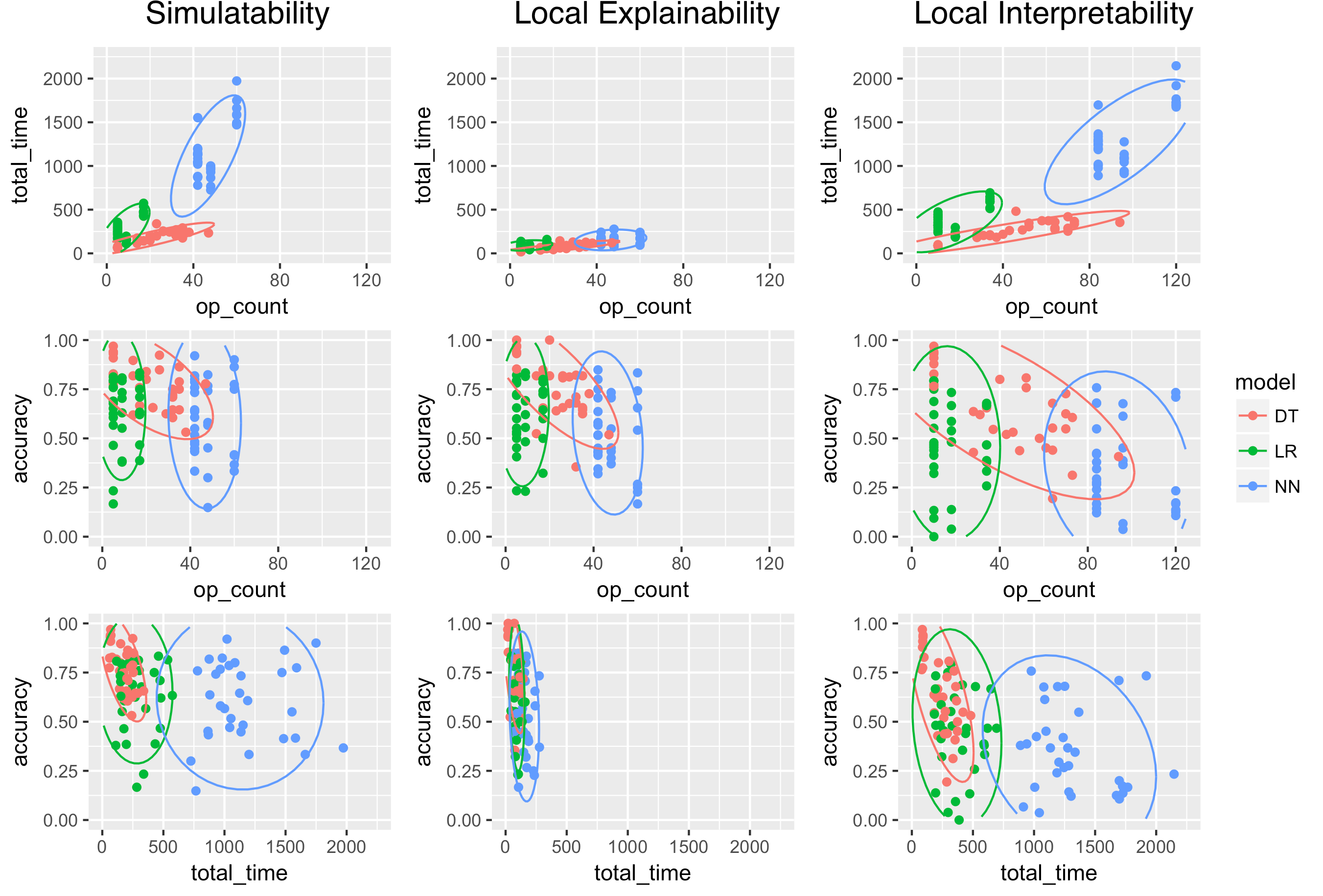}
\caption{Comparisons shown are between total operations for a particular trained model and input, the time taken by the user to complete the task, and the accuracy of the users on that task for the simulatability (original input), \whatif (perturbed input), and the combined local interpretability (getting both tasks correct) tasks.  The total time shown is in seconds. The total operation count is for the simulatability task on the specific input; this is the same for both \whatif and simulatability except for in the case of the decision tree models, where operation counts differ based on input.  The local interpretability operation count is the sum for the simulatability and \whatif task operation counts.  Accuracy shown is averaged over all users who were given the same input for that task and trained model.  The models considered are decision trees (DT), logistic regression models (LR), and neural networks (NN). The ellipses surrounding each group depict the covariance between the two displayed variables, and capture 95\% of the sample variance.}
\label{fig:timeVaccVops}
\end{center}
\end{figure*}

In order to evaluate our preregistered hypotheses, we considered the relationship between total operation counts, time, and accuracy on the simulatability, \whatif, and combined local interpretability tasks.  The graphs showing these relationships, including ellipses that depict the degree to which the different measurements are linearly related to each other, are shown in Figure \ref{fig:timeVaccVops}.  The time and accuracy given for the simulatability and \whatif tasks are separated individually for those tasks in the first two columns of the figure, while the final local interpretability column includes the sum of the time taken by the user on both tasks and credits the user with an accurate answer only if both the simulatability and \whatif tasks were correctly answered.  The accuracies as displayed in the figure are averaged over all users given the same input into the trained model.  All total operation counts given are for the simulation task on the specific input.  In the case of the \whatif task for decision trees, this operation count is for the simulatability task on the perturbed input;  the logistic regression and neural network simulatability operation counts do not vary based on input.  The local interpretability total operation count is the sum of the counts on the simulatability and \whatif tasks.  Additionally, we considered the effect on time and accuracy of just the arithmetic operation counts.  The overall trends are discussed below.

\subsection{Assessing the Relationship Between Runtime Operations and Time}

\subsubsection{The number of operations has a positive relationship with the time taken.}
Across all three interpretability tasks it appears clear that as the number of operations increases, the total time taken by the user also increases (see the first row of Figure \ref{fig:timeVaccVops}).  This trend is especially clear for the simulatability task, validating Hypothesis 1.  This effect is perhaps not surprising, since the operation count considered is for the simulatability task and the representations given focus on performing each operation.  

\subsubsection{Users were locally interpreting the \whatif task.}
Users spent much less time on the local explainability task than the simulatability task across all models.  The difference suggests that users were actually locally interpreting the model on the \whatif task as opposed to re-simulating the whole model.  

\subsubsection{The time taken to simulate neural networks might not be feasible in practice.}
The neural network simulation time was noticeably greater than that of the decision tree and logistic regression.  In some cases, the time expended was greater than 30 minutes.  A user attempting the simulate the results of a model might give up or be unable to dedicate that much time to the task.  The study takers likely feared lack of compensation if they gave up.  This result suggests that in time constrained situations, neural networks are not simulatable. 

\subsection{Assessing the Relationship Between Runtime Operations and Accuracy}

\subsubsection{The relationship between accuracy and operation count is clear for decision trees but not the other model types.}
As the total number of runtime operations increases, we hypothesized that the accuracy would decrease.  In the second row of Figure \ref{fig:timeVaccVops} we can see that this trend appears to hold clearly for all three interpretability tasks for the decision tree models, but there is no clear trend for the logistic regression and neural network models.  This lack of effect may be due to the comparatively smaller range of operation counts examined for these two model types, or it may be that the local interpretability of these model types is not as related to operation count as it is for decision trees.  The lack of overlap in the ranges for the operation counts of logistic regression and neural networks also makes it hard to separate the effects of the model type on the results.  

\subsubsection{Some users might not have understood the logistic regression and neural network tasks.}
Because the logistic regression and neural network tasks could be considered more challenging than the decision tree task, there may have been noise introduced by the variability in user ability to perform the task.  While operation counts might influence the accuracy for users who are able to understand the base task, this trend may be hidden by the fact that some users who were confident did not understand the task.  

%
%Our hypothesis that participant accuracy would decrease as the operation count increased was based on the idea that each operation performed gave the user a chance to make a mistake, and so more operations would increase the chance of mistakes, thus decreasing the accuracy.  In order to test this more directly, we considered the intermediate steps calculated by users as part of the logistic regression and neural network models and counted the number of mistakes made in the simulation of these models.  Mistakes were counted only the first time they were made, so that an early mistake was penalized only once and future calculations were made with respect to that entry.  The results, in Figure \ref{fig:opsVerr}, seem to confirm that as users calculate more operations they make more mistakes, though again the effect is hard to separate from the model type.
%
%\begin{figure}[htbp]
%\begin{center}
%\includegraphics[width=3in]{figures/totalopsVmistakes_nopresentationerrors.png}
%\caption{The total operation count versus the number of mistakes each user makes.  A mistake is defined as entering the incorrect value given the users previously entered values.  So, a mistake is counted only once. }
%\label{fig:opsVmistakes}
%\end{center}
%\end{figure}
%
%

%As can be seen in the bottom row of Figure \ref{fig:timeVaccVops}, user accuracy on the tasks varies somewhat with time.  Although the accuracy seems in general to decrease as users take more time on the task, it does not appear to be a strong trend except for with decision tree models.

\section{Discussion and Conclusion}
We investigated the local interpretability of three common model types: decision trees, logistic regression, and neural networks, and our user study provides evidence for the folk hypotheses that decision trees and logistic regression models are locally interpretable, while neural networks are not.
We also found that decision trees are more locally interpretable than logistic regression or neural network models.
We also showed that as the number of runtime operations increase, participants take longer to locally interpret a model, and they become less accurate on local interpretation tasks.
This runtime operations metric provides some insight into the local interpretability of the discussed models and representations, and could indicate to practitioners the extent to which their models fulfill a lower bound requirement of interpretability.
Further work is needed to consider the extent to which the metric generalizes to other model types.
In addition, we found that users were consistently unable to locally interpret the largest operation count neural networks shown to them, and their inability to simulate such neural networks could suggest that users struggle to locally interpret models more than 100 operations.
Because we did not give users other models of similar operation count due to their potential display size to the user, further work is needed to verify if users inability to locally interpret large neural networks was caused by the number of operation counts or neural networks themselves. 

Further, there are many caveats and limitations to the reach of this work.
The domain-agnostic nature of our synthetic dataset has transferability advantages, but also has disadvantages in that it does not study interpretability within its target domain.
The definitions of local interpretability that we assess here --- simulatability and \whatif --- are limited in their reach and the specific user study setup that we introduce may be limited in capturing the nuance of these definitions.
Still, this work provides a starting point for designing user studies to validate notions of interpretability in machine learning.
Such controlled studies are delicate and time-consuming, but are ultimately necessary in order for the field to make progress.

\bibliography{interpretability}

\begin{thebibliography}{}

\bibitem[\protect\citeauthoryear{Adler \bgroup et al\mbox.\egroup
  }{2018}]{adler2018auditing}
Adler, P.; Falk, C.; Friedler, S.~A.; Nix, T.; Rybeck, G.; Scheidegger, C.;
  Smith, B.; and Venkatasubramanian, S.
\newblock 2018.
\newblock Auditing black-box models for indirect influence.
\newblock {\em Knowledge and Information Systems} 54(1):95--122.

\bibitem[\protect\citeauthoryear{Allahyari and
  Lavesson}{2011}]{Allahyari2011understandability}
Allahyari, H., and Lavesson, N.
\newblock 2011.
\newblock User-oriented assessment of classification model understandability.
\newblock In {\em 11th scandinavian conference on Artificial intelligence}.
\newblock IOS Press.

\bibitem[\protect\citeauthoryear{Breiman}{2001}]{breiman2001random}
Breiman, L.
\newblock 2001.
\newblock Random forests.
\newblock {\em Machine learning} 45(1):5--32.

\bibitem[\protect\citeauthoryear{Buitinck \bgroup et al\mbox.\egroup
  }{2013}]{sklearn_api}
Buitinck, L.; Louppe, G.; Blondel, M.; Pedregosa, F.; Mueller, A.; Grisel, O.;
  Niculae, V.; Prettenhofer, P.; Gramfort, A.; Grobler, J.; Layton, R.;
  VanderPlas, J.; Joly, A.; Holt, B.; and Varoquaux, G.
\newblock 2013.
\newblock {API} design for machine learning software: experiences from the
  scikit-learn project.
\newblock In {\em ECML PKDD Workshop: Languages for Data Mining and Machine
  Learning},  108--122.

\bibitem[\protect\citeauthoryear{Datta, Sen, and
  Zick}{2016}]{datta2016algorithmic}
Datta, A.; Sen, S.; and Zick, Y.
\newblock 2016.
\newblock Algorithmic transparency via quantitative input influence: Theory and
  experiments with learning systems.
\newblock In {\em Security and Privacy (SP), 2016 IEEE Symposium on},
  598--617.
\newblock IEEE.

\bibitem[\protect\citeauthoryear{Doshi-Velez and Kim}{2017}]{doshi2017towards}
Doshi-Velez, F., and Kim, B.
\newblock 2017.
\newblock Towards a rigorous science of interpretable machine learning.
\newblock {\em arXiv preprint arXiv:1702.08608}.

\bibitem[\protect\citeauthoryear{Goodman and
  Flaxman}{2016}]{goodman2016european}
Goodman, B., and Flaxman, S.
\newblock 2016.
\newblock European union regulations on algorithmic decision-making and a"
  right to explanation".
\newblock presented at the 2016 ICML Workshop on Human Interpretability in
  Machine Learning (WHI 2016), New York, NY.

\bibitem[\protect\citeauthoryear{Guidotti \bgroup et al\mbox.\egroup
  }{2018}]{guidotti2018survey}
Guidotti, R.; Monreale, A.; Ruggieri, S.; Turini, F.; Giannotti, F.; and
  Pedreschi, D.
\newblock 2018.
\newblock A survey of methods for explaining black box models.
\newblock {\em ACM Computing Surveys (CSUR)} 51(5):93.

\bibitem[\protect\citeauthoryear{Henelius \bgroup et al\mbox.\egroup
  }{2014}]{henelius2014peek}
Henelius, A.; Puolam{\"a}ki, K.; Bostr{\"o}m, H.; Asker, L.; and Papapetrou, P.
\newblock 2014.
\newblock A peek into the black box: exploring classifiers by randomization.
\newblock {\em Data mining and knowledge discovery} 28(5-6):1503--1529.

\bibitem[\protect\citeauthoryear{Ike-Nwosu}{2018}]{vmbook}
Ike-Nwosu, O.
\newblock 2018.
\newblock {\em Inside Python Virtual Machine}.
\newblock Lean publishing, 1st edition.
\newblock chapter 5 Code Objects,  68--78.

\bibitem[\protect\citeauthoryear{Lage \bgroup et al\mbox.\egroup
  }{2018a}]{lage2018evaluation}
Lage, I.; Chen, E.; He, J.; Narayanan, M.; Gershman, S.; Kim, B.; and
  Doshi-Velez, F.
\newblock 2018a.
\newblock An evaluation of the human-interpretability of explanation.
\newblock In {\em Conference on Neural Information Processing Systems (NeurIPS)
  Workshop on Correcting and Critiquing Trends in Machine Learning}.

\bibitem[\protect\citeauthoryear{Lage \bgroup et al\mbox.\egroup
  }{2018b}]{lage2018humaninloop}
Lage, I.; Slavin~Ross, A.; Kim, B.; J.~Gershman, S.; and Doshi-Velez, F.
\newblock 2018b.
\newblock Human-in-the-loop interpretability prior.
\newblock In {\em Conference on Neural Information Processing Systems
  (NeurIPS)}.

\bibitem[\protect\citeauthoryear{Lakkaraju, Bach, and
  Leskovec}{2016}]{lakkaraju2016}
Lakkaraju, H.; Bach, S.~H.; and Leskovec, J.
\newblock 2016.
\newblock Interpretable decision sets: {A} joint framework for description and
  prediction.
\newblock In {\em ACM SIGKDD Conference on Knowledge Discovery and Data Mining
  (KDD)}.

\bibitem[\protect\citeauthoryear{Lipton}{2018}]{lipton2018mythos}
Lipton, Z.~C.
\newblock 2018.
\newblock The mythos of model interpretability.
\newblock {\em Queue} 16(3):30.

\bibitem[\protect\citeauthoryear{McCabe}{1976}]{mccabe1976complexity}
McCabe, T.~J.
\newblock 1976.
\newblock A complexity measure.
\newblock {\em IEEE Transactions on software Engineering} (4):308--320.

\bibitem[\protect\citeauthoryear{Molnar}{2018}]{molnar2018interpretable}
Molnar, C.
\newblock 2018.
\newblock {\em Interpretable Machine Learning}.
\newblock https://christophm.github.io/interpretable-ml-book/.
\newblock \url{https://christophm.github.io/interpretable-ml-book/}.

\bibitem[\protect\citeauthoryear{Ned}{2008a}]{pycfile}
Ned, B.
\newblock 2008a.
\newblock The structure of .pyc files.
\newblock Blog.
\newblock
  \url{https://nedbatchelder.com/blog/200804/the_structure_of_pyc_files.html}.

\bibitem[\protect\citeauthoryear{Ned}{2008b}]{pychack}
Ned, B.
\newblock 2008b.
\newblock Wicked hack: Python bytecode tracing.
\newblock Blog.
\newblock
  \url{https://nedbatchelder.com/blog/200804/wicked_hack_python_bytecode_tracing.html}.

\bibitem[\protect\citeauthoryear{Olah \bgroup et al\mbox.\egroup
  }{2018}]{olah2018the}
Olah, C.; Satyanarayan, A.; Johnson, I.; Carter, S.; Schubert, L.; Ye, K.; and
  Mordvintsev, A.
\newblock 2018.
\newblock The building blocks of interpretability.
\newblock {\em Distill}.
\newblock https://distill.pub/2018/building-blocks.

\bibitem[\protect\citeauthoryear{Poursabzi-Sangdeh \bgroup et al\mbox.\egroup
  }{2017}]{poursabzi2017manipulating}
Poursabzi-Sangdeh, F.; Goldstein, D.~G.; Hofman, J.~M.; Vaughan, J.~W.; and
  Wallach, H.
\newblock 2017.
\newblock Manipulating and measuring model interpretability.
\newblock Transparent and Interpretable Machine Learning in Safety Critical
  Environments Workshop at NIPS.

\bibitem[\protect\citeauthoryear{Prolific}{2014}]{prolific}
Prolific.
\newblock 2014.
\newblock https://prolific.ac/, last accessed on June 5th, 2019.

\bibitem[\protect\citeauthoryear{Ribeiro, Singh, and
  Guestrin}{2016}]{ribeiro2016should}
Ribeiro, M.~T.; Singh, S.; and Guestrin, C.
\newblock 2016.
\newblock Why should i trust you?: Explaining the predictions of any
  classifier.
\newblock In {\em Proceedings of the 22nd ACM SIGKDD international conference
  on knowledge discovery and data mining},  1135--1144.
\newblock ACM.

\bibitem[\protect\citeauthoryear{Selbst and
  Barocas}{2018}]{selbst2018intuitive}
Selbst, A.~D., and Barocas, S.
\newblock 2018.
\newblock The intuitive appeal of explainable machines.
\newblock {\em Fordham Law Review. Forthcoming. Available at SSRN:
  \url{https://ssrn.com/abstract=3126971}}.

\bibitem[\protect\citeauthoryear{Selbst and
  Powles}{2017}]{selbst2017meaningful}
Selbst, A.~D., and Powles, J.
\newblock 2017.
\newblock Meaningful information and the right to explanation.
\newblock {\em International Data Privacy Law} 7(4):233--242.

\bibitem[\protect\citeauthoryear{Ustun and Rudin}{2016}]{ustun2016supersparse}
Ustun, B., and Rudin, C.
\newblock 2016.
\newblock Supersparse linear integer models for optimized medical scoring
  systems.
\newblock {\em Machine Learning} 102(3):349--391.

\bibitem[\protect\citeauthoryear{Wachter, Mittelstadt, and
  Floridi}{2017}]{wachter2017right}
Wachter, S.; Mittelstadt, B.; and Floridi, L.
\newblock 2017.
\newblock Why a right to explanation of automated decision-making does not
  exist in the general data protection regulation.
\newblock {\em International Data Privacy Law} 7(2):76--99.

\end{thebibliography}
\bibliographystyle{aaai}
\end{document}